# Over-Sampling in a Deep Neural Network

Andrew J.R. Simpson [#1]

[#] *Centre for Vision, Speech and Signal Processing, Surrey University*
*Surrey, UK*
[1] `Andrew.Simpson@Surrey.ac.uk`

*Abstract*—**Deep neural networks (DNN) are the state of the art on many engineering problems such as computer vision and audition. A key factor in the success of the DNN is scalability – *bigger networks work better*. However, the reason for this scalability is not yet well understood. Here, we interpret the DNN as a discrete system, of linear filters followed by nonlinear activations, that is subject to the laws of sampling theory. In this context, we demonstrate that over-sampled networks are more selective, learn faster and learn more robustly. Our findings may ultimately generalize to the human brain.**

## I. INTRODUCTION

The simplest form of feed-forward deep neural network (DNN) features at least two layers of neurons that, via a full complement of weights, are fully connected to inputs from the layer below [1], [2]. At each neuron, the incoming data are multiplied by the respective weights and summed. This can be interpreted as a linear filter. The linear sum is then passed through a nonlinear activation function. Thus, we may interpret the DNN as a cascade of linear filters and layer-wise nonlinear activations.

In this context, the goal of training a DNN is to find the filters whose outputs best elucidate the classification problem in hand. A typical problem for a DNN is image classification [1], whereby some rectangular matrix of pixel intensities is unpacked into a vector which constitutes the input to the first layer of the network. These pixels are sampled from some underlying continuous image signal. Therefore, not only is the network itself discrete by definition but the input data are discretely sampled at a corresponding rate (the input layer is the same length as the data).

If we view the DNN as a discrete system with a fixed sampling rate, at which the filters and data are represented, then it follows that the Nyquist limits apply to the data, the filters and the intermediate representations. The use of nonlinear activation functions in a neural network results in both harmonic and intermodulation distortion. Products of either class of distortion may legitimately be represented, and hence exploited by a network, where it falls within the Nyquist limit. However, conversely, products of either class of distortion may fall beyond the Nyquist limit and hence end up 'folded down' (aliased) into the Nyquist range.

Following this logic, wider networks provide the following obvious advantages; 1) greater bandwidth of input data may be represented, 2) higher-order filters may be learned, and 3) less aliasing of high order distortion products. This leads to the question; can we improve the performance of a DNN simply by over-sampling without obtaining higher resolution data? In this paper, we demonstrate that over-sampling a DNN can both improve performance and mitigate over-fitting within the context of a typical computer vision problem.

## II. METHOD

We chose the well-known computer vision problem of hand-written character classification using the MNIST dataset [3]. At an image resolution of 28x28 pixels, this problem has been thoroughly solved using DNN [1], [3], [4]. So, to provide some headroom, we unpacked the 28x28 pixels into vectors of 784 (28x28 = 784) pixels and then decimated the vectors by a factor of 16. This resulted in vectors of length 49 (784/16 = 49), giving an effective image resolution of 7x7 pixels, as illustrated (by re-wrapping the vector into a matrix) for an example decimated digit in Fig. 1. Prior to decimation, pixel intensities were normalized to zero mean.

Consistent with Hinton *et al.* [1], we used sigmoid activation functions $[output = 1/(1 + \exp(-input))]$ in a fully connected network. However, in order to avoid confounding factors of re-sampling between layers (and in contrast to Hinton *et al.* [1]), we kept the first two layers at the same sample rate (i.e., size) as the inputs before folding down to a 10-unit softmax output layer, corresponding to the 10-way classification problem. This gave us a reference network with dimensions 49x49x10 units that serves as a baseline.

By holding the bandwidth of the input constant and over-sampling by different factors, we were able to test networks at different scales without introducing new information. Thus, we were able to identify changes in performance that may only be attributed to features of the architecture. The low-resolution vectors were up-sampled by factors of 2, 4, 8 and 16 using interpolation. For each sampling rate, we built a corresponding model where the first two layers were scaled up accordingly. Thus, the largest model featured 784x784x10 units (49x16=784) but the input contained no more information (bandwidth) than the baseline (49x49x10) model.

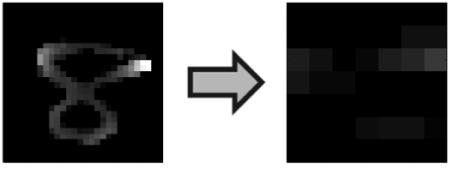

**Fig. 1. Example MNIST image decimated by factor of 16.** We took the 28x28 pixel images, unpacked them into a vector and decimated the vectors by a factor of 16, yielding an effective 7x7 pixel representation (represented here, for illustration, as a matrix re-wrapped from the vector for illustration).

We trained the various networks on the 60,000 training examples from the MNIST dataset [3] using stochastic gradient descent (SGD). Each model was trained with 400 complete sweeps of the entire training dataset and the resulting models were tested on the 10,000 separate test examples at 25, 50, 100, 200 and 400 (full-sweep) iteration points. This allowed us to assess performance of the models, training convergence as a function of iterations, and to look for evidence of over-fitting in terms of non-monotonicity. Models were trained without dropout.

In order to quantify the evolving selectivity of the weights (interpreted as filters) as the networks were trained, we also calculated the transfer function ($H$) of each filter (the weights of each neuron taken as linear filter coefficients), at the same step points (25, 50, 100, 200, 400 iterations), and computed a ratio ($g$, in dB) between the maximum and minimum gain in each transfer function;

$$g = 10 log 10 \left(\frac{max(H)}{min(H)}\right) \quad (1)$$

This allowed us to broadly characterize the selectivity of each filter. These measures were then averaged across each layer (for each model). This allowed us to quantify the evolution of overall selectivity in the filters of each model and hence to provide some insight into what the networks were learning during training and during over-fitting.

### III. RESULTS AND DISCUSSION

Fig. 2a plots classification error for each model, applied to classifying the separate test data (10,000 examples), as a function of the number of iterations of training. The 16x over-sampled model (784x784x10 units) performed best overall, converged most rapidly and showed no sign of over-fitting. The models at 1x and 2x over-sampling performed worse overall and did not show signs of convergence sufficient to establish whether they would ultimately provide a non-monotonic error function. The 4x and 8x models were competitive during under-fitting (low numbers of iterations) but later showed extreme non-monotonicity indicating over-fitting. Thus, contrary to conventional expectations, the largest (16x) model showed the least sign of over-fitting, and the 8x model showed less severe over-fitting than the 4x model. Contrary to popular belief, these results demonstrate that bigger models can over-fit *less*. Indeed, it appears that the 16x over-sampled model is not in need of regularisation, despite the large number of parameters in the model (>600,000).

Fig. 2b plots the layer-averaged crest factor (average difference in gain between the maxima and minima of each transfer function) for the neuronal filters of each model at layer 2. Here, we assume that a model which is learning well should refine the selectivity of its filters such that the crest factor increases until convergence. The 1x model shows the smallest crest factor (indicating the least selective filters) and the poorest evolution of selectivity – the crest factor gets smaller as the training proceeds. The higher order models show progressively higher crest factors, indicating more selective filters, but only the 16x over-sampled model shows monotonic increase in average crest factor. The alternative over-sampled models (2x, 4x, 8x) initially become more selective but subsequently get less selective. This profile tends to coincide with the over-fitting profile of Fig. 2a, where, in particular, the models at 4x and 8x show the earliest decrease of selectivity and onset of over-fitting. Across models, the average classification error was inversely correlated with the average crest factor ($r$ = -0.96, $P$ < 0.01, *Pearson Product-Moment Correlation*). Hence, we may interpret improvements of performance as resulting, at least partially, from refinement of filter selectivity. We may also interpret the non-monotonic trends of selectivity to reflect over-fitting and therefore the lack of over-fitting demonstrated (in Fig. 1a) for the 16x model presumably results from continuous refinement of filter selectivity. In this context, the fact that the 1x (not over-sampled) model performs the worst (Fig. 1a) and shows a steadily decreasing selectivity (Fig. 1b), suggests that it already exists in a state of over-fitting by the 25[th] iteration, and hence the main reason for its poor performance is over-fitting. This most likely explains the similar, monotonic function of the 2x model.

To summarize, in the results we see two types of over-fitting in evidence, the first type is the early over-fitting shown in the selectivity function of the 1x model (Fig. 2b) and the second type is the late over-fitting (Fig. 2a) shown in the 4x and 8x models (that is also reflected in the non-monotonic selectivity functions of Fig. 2b). From this we may conclude the following: 1) Learning rate scales with over-sampling rate, and 2) onset of over-fitting is delayed longer at higher degrees of over-sampling. Therefore, the more we over-sample a DNN the faster it will learn and the less it will over-fit.

We conducted the same analysis on the third layer of each model and found that the results were not interpretable.

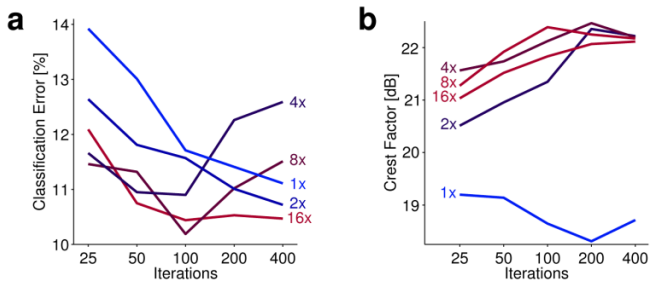

**Fig. 2. Training over-sampled models: evolution of performance and selectivity. a** Classification error in the test set as a function of iterations (each iteration indicates a full sweep of SGD) for different degrees of over-sampling. *Note that the 16x over-sampled model provides the least evidence of over-fitting.* **b** Average crest-factor ($g$ of Eq. 1, in dB) of the transfer functions of neurons in the second layer. *Note that the 16x over-sampled model provides the only monotonic function.*

These results suggest that over-sampling may be a useful alternative to regularization by dropout [5], [6]. The efficacy of dropout is typically interpreted as resulting from reduced co-adaptations to 'rare feature combinations' in input data. However, if the over-fitting in Fig. 2a was the result of rarely occurring feature-coincidences in the data, then over-sampling would not improve matters. Therefore, it may be that the question of over-fitting needs further consideration. In particular, our results suggest that the over-fitting demonstrated here is at least partly the result of aliasing, rather than rare feature coincidences in the data, and hence is mitigated by over-sampling.

Practically speaking, dropout is well known for increasing the training time necessary [5]. By comparison, it would appear that over-sampling improves training time at the expense of increased network size. We do not give a comparison of the two techniques here for the simple reason that we tried the same analysis with 50% dropout but the models performed so poorly that we could not proceed. E.g., the same models with 50% dropout showed test error of around 80-90% after 150 iterations. Hence, it would appear that in our paradigm dropout is redundant at best.

The finding that more refined filters and reduced aliasing lead to better performance is not hard to understand, given that aliasing provides a very abrupt and arbitrary projection of high order distortion products back into the Nyquist band. For example, consider energy near 10x the Nyquist limit. If this energy shifts in frequency a small amount, its aliased location in the Nyquist range shifts a very large amount. Hence, a network that learns aliased relations would likely have poor invariance characteristics and would likely behave badly during training due to the extreme gradients produced by such interactions. So, it may be that over-sampled networks are able to learn greater degrees of invariance, due to the reduced aliasing, and hence this may explain the lack of over-fitting in evidence for the 16x model. Future work might seek to elucidate this question. Finally, it is remarkable that the models perform as well as they do, given the extremely decimated data. This suggests that the low-frequency information in such images accounts for the greatest part of the classification accuracy.

IV. CONCLUSION

In this paper we have provided a sampling theoretic interpretation of deep neural networks, where the system is interpreted in terms of learned filters and management of nonlinear distortion. We have demonstrated that over-sampled networks perform better, are more selective, learn more rapidly and over-fit less. These results have broad implications for how deep neural networks are interpreted, designed and understood.

As a final thought, our findings suggest that a vastly over-sampled system would learn robustly and rapidly. Thus, it may be that our findings generalize to the vast scale of the biological brain.

ACKNOWLEDGMENT

AJRS was supported by grant EP/L027119/1 from the UK Engineering and Physical Sciences Research Council (EPSRC).